\documentclass[11pt]{article}
\usepackage{times}

\usepackage{times}
\usepackage{helvet}
\usepackage{courier}
\usepackage{url}
\usepackage{graphicx}
\usepackage{amsmath,amssymb,amsthm}
\usepackage{booktabs}
\usepackage{xcolor}
\usepackage{algorithm}
\usepackage{algorithmic}

\theoremstyle{definition}
\newtheorem{definition}{Definition}
\newtheorem{assumption}{Assumption}
\newtheorem{remark}{Remark}

\usepackage{tikz}
\usetikzlibrary{shapes.geometric, arrows.meta, positioning, calc, shadows.blur}

\tikzset{
    block/.style={rectangle, draw=blue!80!black, thick, fill=white, minimum width=2.5cm, minimum height=1.1cm, align=center, rounded corners=2pt, font=\sffamily\small, blur shadow},
    process/.style={rectangle, draw=orange!80!black, thick, fill=orange!5, minimum width=2.2cm, minimum height=0.9cm, align=center, font=\sffamily\small},
    signal/.style={ellipse, draw=gray!80, thick, fill=gray!5, font=\sffamily\scriptsize, align=center},
    controller/.style={rectangle, draw=red!80!black, thick, fill=red!5, minimum width=2.2cm, minimum height=0.9cm, align=center, font=\sffamily\small, double},
    arrow/.style={-{Stealth[scale=1.2]}, thick, draw=black!80}
}

\title{Kalman-Inspired Runtime Stability and Recovery in Hybrid Reasoning Systems}

\author{
Barak Or\\
Google and Reichman Tech School \\
Reichman University \\
MetaOr Artificial Intelligence \\
\texttt{barak@metaor.ai}
}
\usepackage[numbers]{natbib}

\begin{document}

\maketitle

\begin{abstract}
Hybrid reasoning systems that combine learned components with model-based inference are increasingly deployed in tool-augmented decision loops, yet their runtime behavior under partial observability and sustained evidence mismatch remains poorly understood. In practice, failures often arise as gradual divergence of internal reasoning dynamics rather than as isolated prediction errors. This work studies runtime stability in hybrid reasoning systems from a Kalman-inspired perspective. We model reasoning as a stochastic inference process driven by an internal innovation signal and introduce cognitive drift as a measurable runtime phenomenon. Stability is defined in terms of detectability, bounded divergence, and recoverability rather than task-level correctness. We propose a runtime stability framework that monitors innovation statistics, detects emerging instability, and triggers recovery-aware control mechanisms. Experiments on multi-step, tool-augmented reasoning tasks demonstrate reliable instability detection prior to task failure and show that recovery, when feasible, re-establishes bounded internal behavior within finite time. These results emphasize runtime stability as a system-level requirement for reliable reasoning under uncertainty.
\end{abstract}

\section{Introduction}

Reasoning systems deployed in real-world environments are increasingly hybrid, combining learned representations with model-based inference, planning, and structured decision-making. Representative examples include learning-augmented planners, neuro-symbolic agents, and LLM-based systems that interact with external tools and execute multi-step decision loops~\cite{garrett2021integrated,besold2021neural,yao2023react}.

Although such systems demonstrate strong performance in controlled benchmarks, their behavior under real-time deployment remains insufficiently characterized. Existing evaluation paradigms primarily emphasize correctness, robustness, or generalization under fixed assumptions~\cite{hendrycks2019benchmarking,gal2016dropout}.
In practice, deployed reasoning agents operate under partial observability, non-stationary environments, and delayed or corrupted feedback. As a result, failures often arise not as isolated prediction errors, but as gradual divergence of internal reasoning dynamics from task-relevant latent processes. Such divergence can accumulate silently and manifest only after substantial degradation in decision quality.

This paper addresses a foundational question: \emph{what constitutes instability in a deployed reasoning system, and how should such instability be detected and managed at runtime?} We define instability as internal inconsistency and unbounded divergence during deployment, rather than semantic incorrectness of individual outputs, and argue that stability must be understood as a dynamic, runtime property of the agent’s inference process.

Inspired by classical estimation theory, we interpret reasoning as a stochastic inference process governed by an internal innovation signal that captures discrepancies between predicted and realized evidence. In Kalman filtering, stability is characterized by bounded error growth, detectability of model mismatch, and recoverability following perturbations~\cite{kalman1960new,anderson2005optimal}. We argue that an analogous notion of stability is necessary for contemporary AI systems whose internal states are high-dimensional and continuously evolving during deployment.

A central motivation for the proposed framework is the separation between learned and model-based components in modern reasoning systems. Learned components excel at semantic generalization but typically lack explicit mechanisms for uncertainty propagation or principled recovery, while model-based components encode structural assumptions that enable reasoning about deviation and correction. Conflating these roles obscures important failure modes in deployed hybrid systems.
From this perspective, instability in reasoning systems should be understood as a system-level phenomenon, rather than solely as a failure of representation or prediction accuracy. A reasoning agent may continue to produce locally coherent outputs while its internal inference dynamics gradually drift away from the task-relevant latent process, a failure mode that is difficult to detect using task-level metrics alone.

We adopt a Kalman-inspired perspective in which stability is characterized by detectability of deviation, bounded divergence, and recoverability, rather than by the absence of error. The relevance of this abstraction lies not in assumptions of linearity or Gaussian noise, but in the explicit separation between prediction, innovation, and correction, which provides a principled basis for monitoring internal consistency during deployment.

In contrast to prior work on robustness, uncertainty estimation, and self-consistency, which primarily targets output correctness or confidence calibration, the present work focuses on the \emph{runtime dynamics} of reasoning under sustained mismatch. We introduce a framework for \emph{Kalman-inspired hybrid reasoning stability} that formalizes reliability in terms of detectability, bounded divergence, and recovery, and operationalizes these properties through measurable innovation-based signals and recovery-time metrics.
\paragraph{Contributions:}
\begin{itemize}
\item We introduce a Kalman-inspired framework for runtime stability in hybrid reasoning systems, modeling reasoning as a stochastic inference process driven by an internal innovation signal.
\item We propose an innovation-based runtime monitoring and recovery mechanism that detects emerging instability and applies recovery actions, including tool fallback, gain modulation, and rollback.
\item We empirically characterize runtime stability under controlled deployment perturbations, demonstrating early instability detection, bounded divergence, and a principled separation between detectable, recoverable, and irrecoverable regimes using a recovery-time metric.
\end{itemize}

The remainder of this paper formalizes the proposed framework and evaluates its behavior under controlled, deployment-like perturbations.
\section{Related Work}

This work relates to hybrid reasoning systems, robustness and uncertainty in AI, and runtime monitoring in deployed agents.

\paragraph{Hybrid reasoning systems.}
Hybrid approaches that combine learned representations with model-based or symbolic reasoning have a long history, including neuro-symbolic systems and learning-augmented planners \cite{besold2021neural,garcez2019,garrett2021integrated}. More recently, large language model (LLM)-based agents integrate neural reasoning with tools, memory, and multi-step interaction loops \cite{yao2023react,schick2023toolformer,wang2023voyager}. Related work on structured prompting and deliberative reasoning includes chain-of-thought and tree-based reasoning methods \cite{wei2022chain,yao2023tree}, as well as iterative self-refinement and reflection-based agents \cite{madaan2023,shinn2023reflexion}. While these approaches improve task-level performance and reasoning quality, existing evaluations largely emphasize correctness or completion, with limited analysis of internal reasoning dynamics over extended deployment horizons.

\paragraph{Robustness and uncertainty.}
Robustness to distribution shift, adversarial perturbations, and uncertainty has been extensively studied, including OOD detection and uncertainty estimation methods \cite{hendrycks2019benchmarking,gal2016dropout}. More recent work highlights that predictive confidence and uncertainty estimates often degrade under realistic dataset shift \cite{ovadia2019,fort2019deep}. In the context of LLMs, self-consistency and agreement-based decoding improve output reliability \cite{wang2023h}, but primarily operate at the output level. These approaches focus on input-output behavior or prediction confidence, and do not directly address sustained internal inconsistency or latent-state drift arising during deployment under semantically valid but mismatched evidence.

\paragraph{Runtime monitoring and control perspectives.}
Runtime monitoring and safety layers have been proposed in autonomous and agentic systems to constrain behavior during deployment \cite{amodei2016concrete,shevlane2023model}. Behavioral testing frameworks argue for evaluation beyond aggregate accuracy, emphasizing systematic failure modes and stress testing \cite{ribeiro2020beyond}. Recent studies of LLM-based agents document recurrent planning and tool-use failures under long-horizon interaction \cite{zhou2024evaluating}. Recovery mechanisms in such systems are typically heuristic and evaluated as binary outcomes. In contrast, classical estimation theory emphasizes innovation-based monitoring, detectability, and bounded stability \cite{kalman1960new,anderson2005optimal}. Inspired by this perspective, our work introduces runtime stability and recovery as measurable system-level properties of deployed reasoning systems, without assuming linear dynamics or probabilistic optimality.

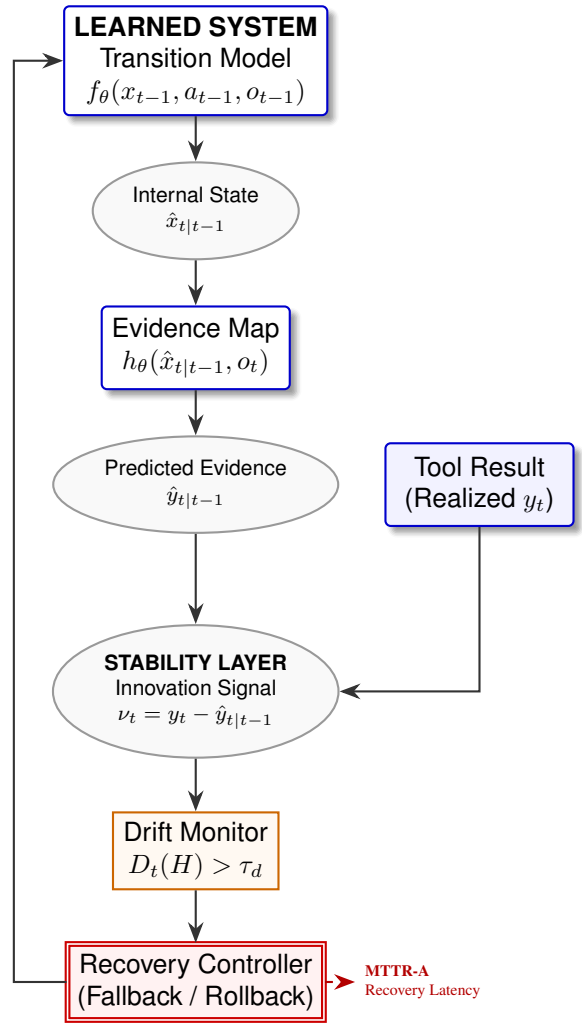
\begin{figure}[!htbp]

\centering
\begin{tikzpicture}[node distance=0.7cm and 0.5cm]

\node [block] (transition) {
    \textbf{LEARNED SYSTEM}\\
    Transition Model\\
    $f_\theta(x_{t-1}, a_{t-1}, o_{t-1})$
};

\node [signal, below=0.6cm of transition] (state) {Internal State\\$\hat{x}_{t|t-1}$};

\node [block, below=0.6cm of state] (evidence_model) {Evidence Map\\$h_\theta(\hat{x}_{t|t-1}, o_t)$};

\node [signal, below=0.6cm of evidence_model] (predicted_y) {Predicted Evidence\\$\hat{y}_{t|t-1}$};

\node [block, right=0.6cm of predicted_y, fill=blue!5] (tool) {Tool Result\\(Realized $y_t$)};

\node [signal, below=1.2cm of predicted_y] (innovation) {
    \textbf{STABILITY LAYER}\\
    Innovation Signal\\
    $\nu_t = y_t - \hat{y}_{t|t-1}$
};

\node [process, below=0.7cm of innovation] (monitor) {Drift Monitor\\$D_t(H) > \tau_d$};

\node [controller, below=0.7cm of monitor] (recovery) {Recovery Controller\\(Fallback / Rollback)};

\node [right=0.4cm of recovery, font=\rmfamily\tiny, color=red!70!black, align=left] (mttr) {\textbf{MTTR-A}\\Recovery Latency};

\draw [arrow] (transition) -- (state);
\draw [arrow] (state) -- (evidence_model);
\draw [arrow] (evidence_model) -- (predicted_y);
\draw [arrow] (predicted_y) -- (innovation);
\draw [arrow] (tool.south) |- (innovation.east);
\draw [arrow] (innovation) -- (monitor);
\draw [arrow] (monitor) -- (recovery);
\draw [arrow, dashed, color=red!70!black] (recovery) -- (mttr);

\draw [arrow] (recovery.west) -- ++(-0.7,0) |- (transition.west);

\end{tikzpicture}
\caption{Hybrid reasoning stability framework. The system decouples learned inference, governed by state transition $f_{\theta}$ and 
evidence mapping $h_{\theta}$ from a Kalman-inspired stability layer that monitors the internal innovation signal $\nu_{t}$
Cognitive drift $D_{t}(H)$ is detected when innovation energy exceeds calibrated 
thresholds, triggering a recovery controller to mitigate unbounded divergence 
through fallback and rollback mechanisms.}
\label{fig:framework}
\end{figure}

\section{Problem Definition}
\label{sec:problem}
We formalize runtime stability in deployed hybrid reasoning systems as a property of their internal inference dynamics under sustained uncertainty.
Our focus is not on task-level correctness or generalization, but on the behavior of a reasoning agent’s internal state during multi-step deployment, where evidence may be delayed, mismatched, or partially inconsistent with the agent’s current intent.
Figure~\ref{fig:framework} provides a schematic overview of the proposed hybrid reasoning stability framework, illustrating the separation between learned state evolution, innovation-based runtime monitoring, and recovery-aware control during deployment.

\subsection{Hybrid Reasoning as Stochastic State Evolution}

We consider a deployed reasoning agent operating over discrete time steps $t$.
The agent maintains an internal reasoning state $x_t \in \mathbb{R}^d$ that summarizes task-relevant latent information, intermediate beliefs, plans, or internal hypotheses.
The agent receives observations $o_t \in \mathcal{O}$ and produces actions $a_t \in \mathcal{A}$.
We assume a hybrid architecture in which state evolution is governed jointly by model-based and learned components. In this formulation, the model-based component provides a structured notion of state evolution and consistency over time, while learned components instantiate the transition and observation mappings. In deployed settings, the agent has access at runtime to observable signals, which we refer to as \emph{evidence}, reflecting the outcome or internal consistency of its reasoning rather than ground-truth correctness.

We represent the agent’s predicted internal state dynamics by a (possibly learned or implicit) transition operator
\begin{equation}
\hat{x}_{t|t-1} = f_\theta(x_{t-1}, a_{t-1}, o_{t-1}),
\label{eq:pred_state}
\end{equation}
which induces an expectation about the evidence that should be observed if the agent’s internal dynamics remain consistent over time.

We therefore introduce an internal observation (or self-evaluation) mapping that produces an expected evidence signal:
\begin{equation}
\hat{y}_{t|t-1} = h_\theta(\hat{x}_{t|t-1}, o_t),
\label{eq:pred_obs}
\end{equation}
where $\hat{y}_{t|t-1}$ represents the agent’s internally predicted evidence under its current reasoning state, rather than a ground-truth observation. 
At runtime, the agent receives an observed evidence vector $y_t \in \mathbb{R}^m$, corresponding to a realization of the evidence-generating process.
The discrepancy between observed evidence $y_t$ and expected evidence $\hat{y}_{t|t-1}$ enables innovation-based consistency monitoring during deployment.

To model uncertainty, we treat the predicted internal dynamics and evidence mapping as stochastic with additive Gaussian perturbations:
\begin{align}
x_t &= f_\theta(x_{t-1}, a_{t-1}, o_{t-1}) + w_t, \qquad w_t \sim \mathcal{N}(0, Q), \label{eq:true_state}\\
y_t &= h_\theta(x_t, o_t) + v_t, \qquad\qquad\qquad\;\;\; v_t \sim \mathcal{N}(0, R), \label{eq:true_obs}
\end{align}
where $Q \in \mathbb{R}^{d \times d}$ and $R \in \mathbb{R}^{m \times m}$ are positive semi-definite covariance matrices, aim to capture epistemic and aleatoric uncertainty arising from partial observability, distribution shift, approximation error, and stochasticity in external tools or environment feedback. They are not assumed to be known \emph{a priori} and are not required to accurately model the true underlying noise.
In this work, they serve as modeling abstractions that capture the relative scale of uncertainty in internal state evolution and observed evidence and may be specified using simple diagonal forms, estimated empirically from nominal trajectories, or absorbed into the innovation scaling used for runtime monitoring.

\subsection{Innovation Process}

\begin{definition}[Innovation]
Let $\hat{y}_{t|t-1}$ denote the predicted evidence under the agent’s internal reasoning state at time $t$, and let $y_t$ denote the realized evidence.  
The runtime innovation is defined as
\begin{equation}
  \nu_t = y_t - \hat{y}_{t|t-1}.  
\end{equation}
\end{definition}

The innovation $\nu_t$ serves as the primary online signal indicating whether the agent’s internal reasoning dynamics remain consistent with incoming evidence.
In classical filtering, the innovation plays a central role in detectability and stability analysis; here, it fulfills an analogous function for hybrid reasoning systems, providing a measurable residual that can be monitored during deployment.
In many systems, the innovation is scalar or heterogeneous (e.g., tool-return residuals or self-consistency discrepancies).
We therefore introduce a scalar innovation energy for monitoring:
\begin{equation}
e_t = \frac{\nu_t^2}{S},
\label{eq:innov_energy}
\end{equation}
where the normalization constant $S$ is estimated from nominal trajectories as the empirical variance of the innovation signal:
\begin{equation}
S 
= \mathbb{E}_{\mathcal{R}_0}\!\left[(\nu_t - \mu_\nu)^2\right] + \varepsilon,
\qquad
\mu_\nu = \mathbb{E}_{\mathcal{R}_0}[\nu_t],
\end{equation}
with $\varepsilon > 0$ a small stabilizing constant.
The value of $S$ is computed once during a nominal calibration phase and held fixed across all experimental conditions.

\subsection{Cognitive Drift}

\begin{definition}[Cognitive Drift]
Let $\{e_t\}_{t \ge 0}$ denote the innovation energy sequence.
For a window size $H \in \mathbb{N}$, define the drift score
\begin{equation}
D_t(H) = \frac{1}{H}\sum_{k=0}^{H-1} e_{t-k}.
\label{eq:def_drift_score}
\end{equation}
Cognitive drift is said to occur at time $t$ if $D_t(H)$ exceeds a calibrated threshold.
\end{definition}

\begin{assumption}[Nominal Innovation Regime]
There exists a nominal operating regime $\mathcal{R}_0$ under which the innovation energy satisfies
\begin{equation}
\mathbb{E}[e_t \mid \mathcal{R}_0] \leq \tau,
\end{equation}
for some finite threshold $\tau>0$.
\end{assumption}

\begin{remark}
Cognitive drift is defined purely in terms of innovation statistics and does not require access to ground-truth latent states or task-level correctness signals.
\end{remark}

\subsection{Semantic Drift}

While innovation-based drift captures inconsistency between predicted and realized evidence, it does not directly reflect deviation from the original task intent.
We therefore define \emph{semantic drift} as the deviation of the internal reasoning state from its initial task representation.
Operationally, semantic drift is measured as
\begin{equation}
s_t = 1 - \cos(x_t, x_0),
\end{equation}
where $x_0$ denotes the initial internal state associated with the task prompt or query. Semantic drift is not used for control, but serves as an evaluation signal to distinguish apparent stability due to adaptation from stability that preserves task intent.

\subsection{Runtime Stability}
\begin{definition}[Runtime Stability]
A reasoning system is runtime stable if there exist constants $B>0$, $B_d>0$, and $\delta\in(0,1)$ such that, for all $t$,
\begin{equation}
\Pr\!\left(e_t \le B\right) \ge 1-\delta,
\qquad
\Pr\!\left(D_t(H) \le B_d\right) \ge 1-\delta.
\end{equation}
\end{definition}

\begin{assumption}[Finite-Horizon Monitoring]
Runtime stability is assessed over finite horizons using sliding-window statistics, rather than asymptotic guarantees.
\end{assumption}

\begin{remark}
Runtime stability does not imply semantic correctness. Stability characterizes bounded internal behavior, not convergence to a correct task solution.
\end{remark}

We define runtime stability in terms of bounded divergence of internal dynamics and bounded innovation under deployment conditions.
Since the internal state $x_t$ may not be directly observable or interpretable, stability is formulated through two coupled conditions: innovation boundedness and bounded drift.

\subsection{Recovery and MTTR-A}
\label{sec:recovery}
We consider reasoning agents equipped with a recovery mechanism that is activated upon drift detection.

\begin{definition}[Recovery Time]
Let $t_0$ denote the first time at which cognitive drift is detected, i.e., $D_{t_0}(H) > \tau_d$.
The recovery time $t_r$ is defined as the earliest time at which the system re-enters a bounded-stability regime:
\begin{equation}
t_r = \inf\{t \geq t_0 : D_t(H) \leq \kappa \tau_d \},
\label{eq:recovery_time}
\end{equation}
where $\kappa \in (0,1)$ is a bounded-stability factor.
\end{definition}

\begin{assumption}[Limited Recoverability]
Recovery is not assumed to be universally feasible under persistent evidence mismatch.
\end{assumption}

\begin{remark}
Detectability and recoverability are distinct properties. A system may reliably detect instability while being unable to recover to a bounded-stability regime.
\end{remark}

A recovery policy $\pi_R$ may include actions such as replanning, rollback to a prior stable state, tool re-querying, model switching, or uncertainty escalation. To quantify recovery efficiency, we adopt the \emph{Mean Time to Recovery for Agentic Systems (MTTR-A)}~\cite{or2025mttr}, defined as
\begin{equation}
\mathrm{MTTR\mbox{-}A} = \mathbb{E}\left[t_r - t_0\right],
\end{equation}
where the expectation is taken over successfully recovered drift events.

\section{Experimental Methodology}
This section describes how the theoretical definitions introduced in Section~\ref{sec:problem} are instantiated and evaluated experimentally. We translate innovation, cognitive drift, runtime stability, and recovery into measurable quantities by analyzing the runtime behavior of reasoning agents under controlled deployment-like conditions. Agents are evaluated over extended horizons with calibrated nominal, perturbation, and recovery phases, enabling direct measurement of drift detection, recovery behavior, and MTTR-A independently of task-level success.

\subsection{Experimental Protocol}
All experiments follow a common three-phase protocol (nominal, perturbation, recovery).

In the \emph{nominal phase}, agents operate under calibrated conditions to collect baseline innovation statistics and derive nominal thresholds for drift detection and stability bounds.

In the \emph{perturbation phase}, partial task solvability is preserved while gradual internal inconsistency is induced, allowing cognitive drift to emerge as a runtime phenomenon.

In the \emph{recovery phase}, drift detection operates online and triggers recovery policies once the drift score exceeds the detection threshold. The system is observed until bounded stability is re-established or recovery fails, enabling measurement of detection latency and recovery duration.

\subsection{Algorithm}

Algorithm~\ref{alg:krsc} provides a concrete instantiation of the runtime stability framework defined in Section~2.
At each reasoning step, the controller computes the innovation signal and corresponding innovation energy, maintains a sliding window of recent values, and evaluates the drift score to determine whether cognitive drift has occurred.
Upon drift detection, a recovery policy is invoked, while normal reasoning updates proceed unchanged otherwise. Specifically, the agent’s standard reasoning update is applied whenever no recovery transition is active.

The function $\textsc{UpdateState}(\cdot)$ denotes the agent’s existing reasoning update, which is treated as a black box. The proposed controller operates solely as an external runtime monitoring and recovery layer that observes runtime consistency signals and intervenes only when instability is detected.

The buffer $\mathcal{B}$ stores the most recent $H$ innovation energy values required to compute the drift score $D_t(H)$ and is not part of the agent’s internal reasoning state. Recovery completion is determined according to the bounded-stability criterion defined in Section~\ref{sec:recovery}, parameterized by the bounded-stability factor $\kappa$, enabling direct measurement of recovery latency via MTTR-A.

\begin{algorithm}[t]
\caption{Kalman-Inspired Runtime Stability Controller}
\label{alg:krsc}
\small
\begin{algorithmic}[1]
\STATE \textbf{Input:} window $H$, thresholds $\tau,\tau_d$, scaling $S_t$, recovery policy $\pi_R$
\STATE Initialize state $x_0$, empty buffer $\mathcal{B}$, recovery flag $\mathsf{off}$
\FOR{$t = 1,2,\ldots$}
    \STATE Predict state $\hat{x}_{t|t-1} \leftarrow f_\theta(x_{t-1},a_{t-1},o_{t-1})$
    \STATE Predict evidence $\hat{y}_{t|t-1} \leftarrow h_\theta(\hat{x}_{t|t-1},o_t)$
    \STATE Execute action / tool call, observe evidence $y_t$
    \STATE Innovation $\nu_t \leftarrow y_t - \hat{y}_{t|t-1}$
    \STATE Energy $e_t \leftarrow \nu_t^2 / S$
    \STATE Update buffer $\mathcal{B} \leftarrow \mathcal{B} \cup \{e_t\}$ (keep last $H$)
    \IF{$|\mathcal{B}|=H$}
        \STATE Drift score $D_t \leftarrow \frac{1}{H}\sum_{e\in\mathcal{B}} e$
        \IF{$D_t > \tau_d$ \AND recovery $\mathsf{off}$}
            \STATE Set detection time $t_0 \leftarrow t$
            \STATE $(x_t,\cdot) \leftarrow \pi_R(x_{t-1},o_t,\nu_t,D_t)$
            \STATE recovery $\mathsf{on}$
        \ELSIF{$D_t \leq \kappa \tau_d$ \AND recovery $\mathsf{on}$}
            \STATE Set recovery time $t_r \leftarrow t$
            \STATE Store $(t_0,t_r)$ for MTTR-A
            \STATE recovery $\mathsf{off}$
        \ELSE
            \STATE $x_t \leftarrow \textsc{UpdateState}(x_{t-1},o_t,y_t)$
        \ENDIF
    \ENDIF
\ENDFOR
\end{algorithmic}
\end{algorithm}

\begin{table}[t]
\centering
\footnotesize
\begin{tabular}{ll}
\toprule
\textbf{Parameter} & \textbf{Value} \\
\midrule
Reasoning horizon $T$ & 30 \\
Drift window $H$ & 3 \\
Perturbation step $t^\star$ & 5 \\
Episodes $N$ & 120 \\
\midrule
Innovation scale $S$ & $\mathbb{E}[\nu_t^2]$ (nominal phase)
 \\
Detection threshold $\tau$ & 90th percentile \\
Drift threshold $\tau_d$ & 95th percentile \\
Stability factor $\kappa$ & 0.85 \\
\midrule
Nominal gain $\alpha$ & 0.35 \\
Rollback mixing $\beta$ & 0.20 \\
\bottomrule
\end{tabular}
\caption{Core experimental parameters and calibration settings. Thresholds and normalization are estimated from nominal trajectories.}
\label{tab:params}
\end{table}

\section{Experimental Instantiation and Results}
This section instantiates the proposed runtime stability framework in a concrete hybrid reasoning setting and reports empirical results under controlled, deployment-like conditions. We describe the task setup, evidence representation, perturbation scenarios, and evaluation protocol, and analyze runtime behavior with respect to instability detection, drift, and recovery. 

\subsection{Task, Dataset, and Agent}
We evaluate the framework on a multi-step, tool-augmented reasoning task based on the \emph{HotpotQA} dataset (validation split, distractor setting)~\cite{yang2018hotpotqa}.

Each evaluation instance corresponds to a sequential reasoning episode in which the agent formulates a query, retrieves external textual evidence via a search tool, and updates an internal reasoning state over multiple steps.

The agent operates in an explicit reasoning loop of fixed horizon $T=30$.
At each step $t$, the agent maintains an internal latent reasoning state $x_t \in \mathbb{R}^d$, produces an implicit prediction of expected evidence represented by its current predicted latent state ${\hat x}_{t|t-1}$, executes a retrieval action, and observes realized evidence $y_t$.
This structure enables online computation of innovation signals, drift statistics, and recovery metrics as defined in Section~\ref{sec:problem}.
The agent itself is not optimized for task performance; its role is to serve as a testbed for analyzing runtime stability properties under deployment conditions.
\subsection{Evidence, Perturbations, and Agent Variants}

\paragraph{Evidence representation and innovation:}
The predicted evidence $\hat y_{t|t-1}$ is instantiated as the agent’s current latent reasoning state ${\hat x}_{t|t-1}$, representing its expectation over incoming evidence prior to retrieval.
Retrieved evidence consists of real passages drawn from the HotpotQA context corpus, each encoded into a fixed-dimensional vector using a pretrained sentence embedding model. The realized evidence vector $y_t$ corresponds to the embedding of the retrieved passage at step $t$. Innovation is measured using a cosine-distance proxy,
\begin{equation}
    \nu_t = 1 - \cos(\hat{y}_{t|t-1}, y_t),
\end{equation}

and the drift score $D_t(H)$ is computed as the mean innovation energy over a sliding window of the last $H=3$ steps.
A scalar innovation energy is defined as $e_t = \nu_t^2 / S$, where $S$ is a normalization constant estimated from nominal trajectories.

\paragraph{Perturbation scenario:}
Cognitive drift is induced through a persistent, structured perturbation applied at the tool interface.
Starting at step $t^\star = 5$, the retrieval tool is intentionally misrouted to use a real but incorrect query drawn from another HotpotQA instance, modeling realistic deployment failures such as stale intent routing or upstream system errors.
The perturbation does not inject synthetic noise or corrupt the data distribution; all retrieved evidence remains real and semantically valid, but becomes systematically misaligned with the agent’s task intent.

\paragraph{Agent variants:}
To isolate the effect of innovation monitoring and recovery, we evaluate two agent variants under identical conditions.
The baseline agent operates without recovery mechanisms and adapts continuously to incoming evidence, potentially stabilizing around an incorrect latent process.
The recovery-aware agent incorporates innovation monitoring, drift detection, and a recovery controller including tool fallback, gain modulation, and rollback mechanisms.

\paragraph{Semantic drift and apparent stability:}
Semantic drift is tracked as a diagnostic signal to distinguish apparent stability achieved through adaptation from bounded stability that preserves task intent.

\subsection{Runtime Stability under Persistent Evidence Mismatch}
\label{subsec:results_main}

\paragraph{Recovery behavior and MTTR-A.}
Upon drift detection, the recovery-aware agent activates a stability controller that attempts to re-enter a bounded-stability regime, as defined in Section Section ~\ref{sec:recovery}.
Under persistent evidence mismatch, recovery is feasible only in a small subset of cases.
Across all detected drift events, bounded stability is successfully re-established in $3\%$ of detected drift events.

Conditioned on successful recovery, the mean time to recovery (MTTR-A) is $4.00 \pm 0.82$ reasoning steps.
Although recovery is rare, successful recoveries occur rapidly.
Importantly, the low recovery rate reflects the prevalence of irrecoverable regimes rather than failure of the detection mechanism.

We evaluate the proposed runtime stability framework under a persistent evidence-mismatch perturbation, without any ablation of recovery components.
All results in this subsection correspond to the full controller configuration described in Section~4 and Algorithm~\ref{alg:krsc}, with a fixed reasoning horizon of $T=30$.

\begin{figure}[t]
    \centering
    \includegraphics[width=0.24\textwidth]{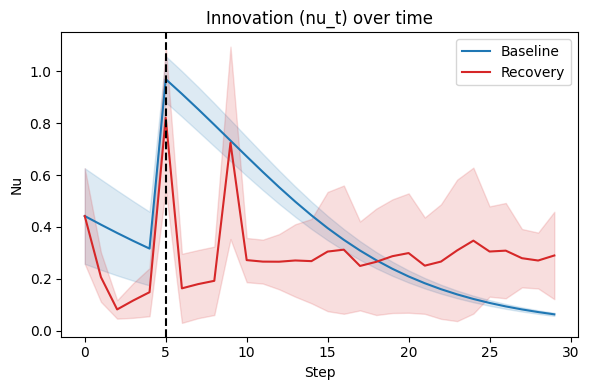}\hfill
    \includegraphics[width=0.24\textwidth]{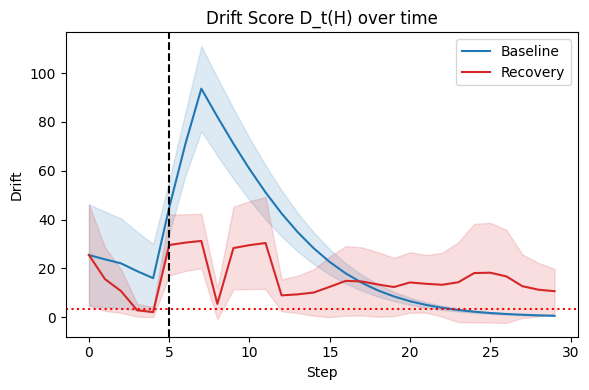}\hfill
    \includegraphics[width=0.24\textwidth]{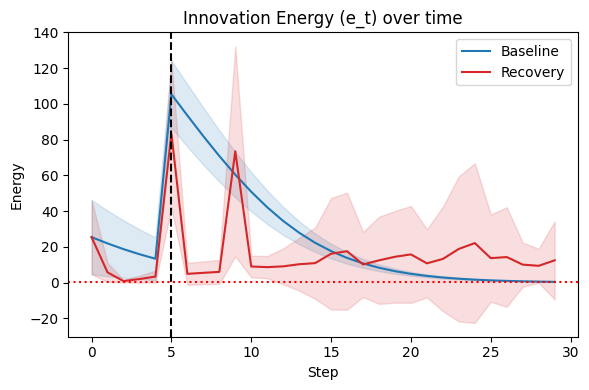}\hfill
    \includegraphics[width=0.24\textwidth]{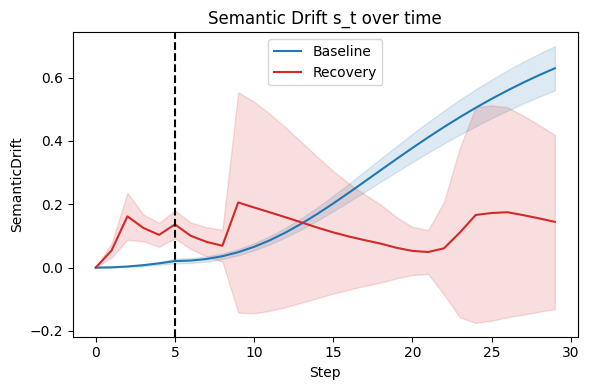}
    \caption{%
    \textbf{Aggregate runtime dynamics under persistent evidence mismatch.}
    From left to right: innovation magnitude $\nu_t$, innovation energy $e_t$, drift score $D_t(H)$, and semantic drift $s_t$.
    Shaded regions denote $\pm 1$ standard deviation across episodes.
    The recovery-aware agent (red) maintains bounded drift and substantially lower semantic deviation than the baseline (blue), despite identical perturbation conditions.
    }
    \label{fig:aggregate_dynamics}
\end{figure}

\paragraph{Aggregate dynamics.}
The aggregate runtime behavior averaged over $N=120$ reasoning episodes is shown below. The perturbation is introduced at step $t^\star=5$ (vertical dashed line). Figure~\ref{fig:aggregate_dynamics} shows the aggregate runtime behavior averaged over episodes.

\begin{figure}[t]
    \centering
    \includegraphics[width=0.24\textwidth]{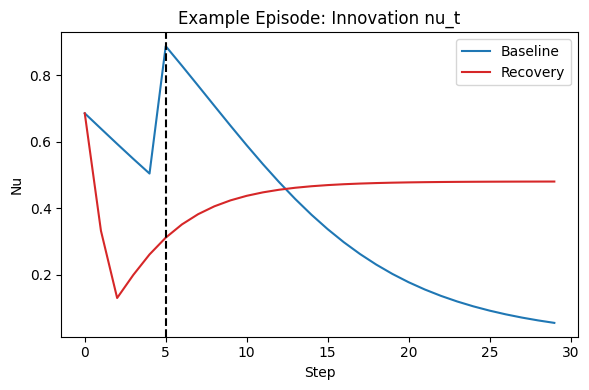}\hfill
    \includegraphics[width=0.24\textwidth]{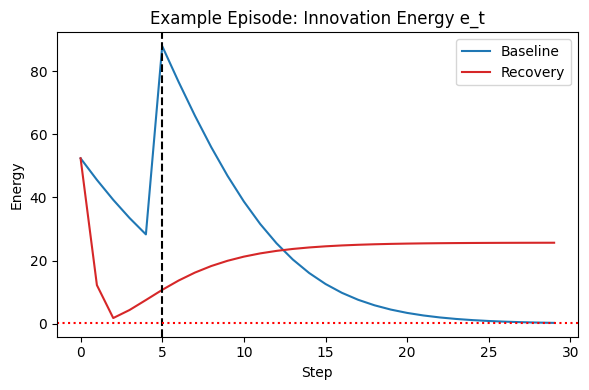}\hfill
    \includegraphics[width=0.24\textwidth]{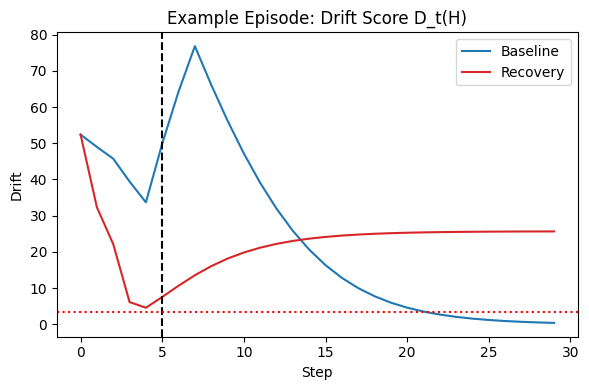}\hfill
    \includegraphics[width=0.24\textwidth]{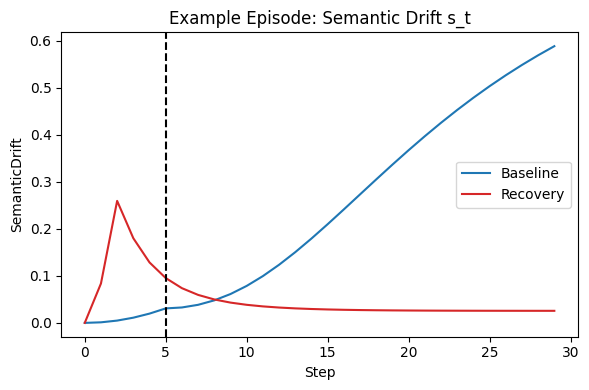}
    \caption{
    \textbf{Representative single-episode trajectories.}
    From left to right: innovation $\nu_t$, innovation energy $e_t$, drift score $D_t(H)$, and semantic drift $s_t$.
    The dashed vertical line marks perturbation onset.
    The baseline converges toward an incorrect latent process, whereas the recovery-aware agent limits semantic deviation despite persistent evidence mismatch.
    }
    \label{fig:single_episode}
\end{figure}

\paragraph{Single-episode behavior:}
Figure~\ref{fig:single_episode} illustrates a representative evaluation episode.
The baseline agent adapts to corrupted evidence by reducing innovation and drift, but accumulates large semantic deviation, converging toward an incorrect latent process.
The recovery-aware agent, in contrast, detects instability, modulates updates, and constrains semantic drift, even when full recovery is not achieved.
This example highlights the distinction between apparent stability through adaptation and bounded stability that preserves task intent.

\begin{figure}[t]
    \centering
    \begin{minipage}[t]{0.52\columnwidth}
        \centering
        \includegraphics[width=\linewidth]{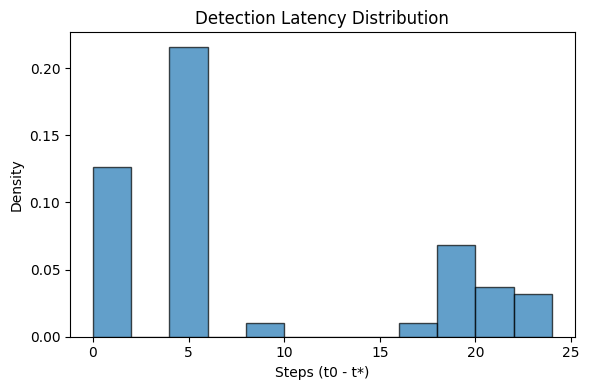}\\
        {\footnotesize (a) Detection latency}
    \end{minipage}
    \hfill
    \begin{minipage}[t]{0.4\columnwidth}
        \centering
        \includegraphics[width=\linewidth]{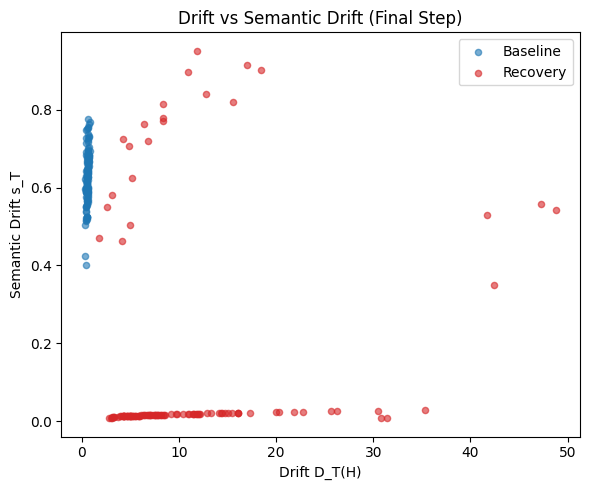}\\
        {\footnotesize (b) Drift vs.\ semantic drift}
    \end{minipage}
    \caption{
    \textbf{Runtime instability diagnostics.}
    (a) Histogram of detection latency $t_0 - t^*$ across detected episodes.
    (b) Drift versus semantic drift at the final reasoning step.
    Each point corresponds to a single episode.
    Recovery-aware agents exhibit bounded drift and lower semantic deviation compared to the baseline.
    }
    \label{fig:runtime_diagnostics}
\end{figure}

\paragraph{Detection latency.}
Detection latency statistics are summarized in Fig.~\ref{fig:runtime_diagnostics}(a).
Cognitive drift is detected in $79.2\%$ of perturbed episodes under the proposed detection criterion, with a mean detection latency of $7.81$ reasoning steps relative to perturbation onset.
Detection consistently precedes overt task failure, confirming that innovation-based monitoring provides an anticipatory signal of emerging runtime instability.

\paragraph{Diagnostic analysis.}
Figure~\ref{fig:runtime_diagnostics}(b) plots the drift score $D_T(H)$ versus semantic drift $s_T$ at the final reasoning step for all episodes.
Baseline agents cluster in regions of high semantic deviation despite reduced drift, reflecting apparent stability achieved through adaptation to corrupted evidence.
In contrast, recovery-aware agents concentrate in a low-drift regime with substantially lower semantic deviation, demonstrating preservation of task intent even when full recovery is not achieved.

Overall, these results validate the core claims of the proposed framework.
Innovation-based monitoring reliably detects runtime instability, drift emerges as a sustained internal phenomenon prior to task failure, and recovery-when feasible-re-establishes bounded internal behavior within finite time.
Crucially, the results reveal a clear separation between detectability and recoverability.

\subsection{Ablation Study}
\label{subsec:ablation}

To assess the contribution of individual components in the proposed runtime stability controller, we conduct an ablation study in which specific recovery mechanisms are selectively disabled.
All variants are evaluated under identical conditions and thresholds, following the experimental protocol described earlier.
Table~\ref{tab:ablation} reports detection rates, recovery rates, and MTTR-A values for each configuration.
\begin{table}[t]
\centering
\footnotesize
\begin{tabular}{lccc}
\toprule
\textbf{Condition} & \textbf{Det.} & \textbf{Rec.} & \textbf{MTTR-A} \\
\midrule
Baseline (no recovery)        & 0.07 & 0.00 & -- \\
Full controller               & 0.79 & 0.03 & $4.00 \pm 1.41$ \\
No rollback                   & 1.00 & 0.11 & $9.62 \pm 7.75$ \\
No gain modulation            & 1.00 & 0.19 & $4.00 \pm 0.00$ \\
No tool fallback              & 0.78 & 0.00 & -- \\
Delayed evidence perturbation & 0.79 & 0.12 & $6.47 \pm 4.01$ \\
\bottomrule
\end{tabular}
\caption{Ablation study of recovery mechanisms.
Detection rate (Det.) denotes the fraction of episodes in which cognitive drift is detected.
Recovery rate (Rec.) denotes the fraction of detected episodes that successfully re-enter a bounded-stability regime.
MTTR-A is reported only for successful recoveries.}
\label{tab:ablation}
\end{table}

The baseline agent exhibits a very low detection rate of $0.07$ and no successful recovery events, confirming that instability often remains silent without explicit innovation-based monitoring.
In contrast, the full controller detects instability in $79\%$ of episodes, but recovers in only $3\%$ of detected cases, with a mean MTTR-A of $4$ steps.
This gap between detection and recovery highlights the prevalence of irrecoverable regimes under persistent evidence corruption.

Disabling rollback increases the recovery rate from $0.03$ to $0.11$, but at the cost of significantly slower and less stable recovery, as reflected by a high mean MTTR-A of $9.62$ steps and large variance.
This indicates that rollback alone can enable recovery but lacks the fine-grained control required for efficient stabilization.

While removing gain modulation increases recovery frequency, it does so by allowing aggressive adaptation, which undermines stability guarantees by allowing aggressive adaptation (recovery rate = 0.19), while maintaining a low and highly consistent recovery time (MTTR-A $= 4$).

Removing tool fallback eliminates recovery entirely (recovery rate $=0$) despite maintaining a high detection rate ($0.78$).
This demonstrates that recovery cannot be achieved through internal state adaptation alone and requires modification of evidence acquisition or tool interaction.

Finally, the delayed-evidence perturbation confirms that the proposed framework generalizes beyond misrouting failures.
Innovation-based detection remains effective ($0.79$ detection rate), while recovery is feasible in a subset of cases ($0.12$ recovery rate) with increased recovery latency (MTTR-A $=6.47 \pm 4.01$).
Together, these results demonstrate a clear separation between detectability and recoverability and emphasize the necessity of explicitly distinguishing detectable, recoverable, and irrecoverable instability regimes in deployed reasoning systems.

\section{Discussion}

We reframe reliability in hybrid reasoning systems as a runtime, system-level property rather than a static consequence of training or offline evaluation. The experimental results show that instability in deployed reasoning agents typically emerges as a gradual internal phenomenon, manifested through sustained innovation abnormalities and semantic drift, rather than as abrupt task failure. This distinction has direct implications for how reasoning systems should be monitored and controlled under real deployment conditions.

A central finding is the clear separation between \emph{detectability} and \emph{recoverability}. Innovation-based monitoring reliably detects emerging instability prior to overt task failure across a range of perturbations. However, recovery is feasible only in a limited subset of cases under persistent evidence mismatch. This gap reflects the prevalence of irrecoverable regimes rather than a weakness of the detection mechanism. In such regimes, continued adaptation to corrupted evidence may reduce local inconsistency while driving the system toward an incorrect latent process, producing apparent stability that masks semantic divergence. Reliable deployment therefore requires not only early detection, but explicit decision-making about when recovery is appropriate and when escalation, rollback, or termination is warranted.

The ablation analysis clarifies the functional role of recovery mechanisms. Tool fallback emerges as a necessary condition for recovery, indicating that stabilization cannot be achieved through internal state adaptation alone under persistent evidence corruption. Gain modulation plays a critical role in regulating the influence of incoming evidence, enabling controlled recovery without aggressive over-adaptation. Rollback mechanisms increase recovery frequency but introduce substantial variance in recovery latency and stability, demonstrating that recovery speed alone is insufficient as a reliability criterion. These results emphasize that recovery is a structured process that must balance responsiveness, robustness, and preservation of task intent.

More broadly, the results challenge evaluation paradigms that equate adaptation with robustness. Baseline agents that continuously adapt to incoming evidence may exhibit reduced innovation and bounded drift while simultaneously accumulating large semantic deviation. Without explicit runtime monitoring, such systems may appear stable under conventional metrics while silently diverging from their intended task or safety envelope. By decoupling internal consistency from task-level success and introducing measurable runtime signals, the proposed framework exposes this failure mode and provides a systems-level notion of bounded stability as resistance to unbounded internal deviation.

The Kalman-inspired formulation should be understood as a conceptual abstraction rather than a probabilistic model of reasoning, and does not rely on linear dynamics, Gaussian noise, or accurate covariance estimation. Its value lies in the explicit separation between prediction, innovation, and correction, which supports principled runtime monitoring and recovery decisions under sustained uncertainty.

\section{Limitations}

The proposed framework is a conceptual abstraction inspired by estimation theory rather than a probabilistic model of reasoning, and does not assume linear dynamics, Gaussian noise, or known covariances. Alternative formulations of innovation or drift may therefore be more appropriate for specific architectures or deployment settings.

The experimental evaluation is restricted to a single class of multi-step, tool-augmented reasoning tasks based on HotpotQA and to sustained evidence-mismatch perturbations. The evaluated agent is not optimized for task-level accuracy and serves as a controlled testbed for analyzing runtime stability; recovery rates and MTTR-A values should thus be interpreted as empirical characterizations under constrained conditions.

Finally, recovery is not guaranteed under persistent evidence corruption, and semantic drift is measured using a diagnostic proxy. The framework assumes access to runtime consistency signals that may be noisy or partially observable in real-world deployments.

\section{Conclusion}

This paper examined runtime stability in hybrid reasoning systems and argued that reliability in deployed agents is fundamentally a dynamic, system-level property rather than a static consequence of training or offline evaluation. By interpreting reasoning as a stochastic inference process driven by an internal innovation signal, we introduced a Kalman-inspired framework that formalizes instability, bounded divergence, and recovery in operational terms.

The proposed framework enables explicit monitoring of internal consistency, early detection of emerging instability, and recovery-aware control during deployment. Experiments on multi-step, tool-augmented reasoning tasks show that innovation-based monitoring reliably detects instability prior to task failure, while recovery-when feasible-re-establishes bounded internal behavior within finite time. Crucially, the results highlight a clear separation between detectability and recoverability, demonstrating that early detection does not guarantee recovery under persistent evidence mismatch.

More broadly, this work emphasizes the importance of runtime monitoring and recovery awareness for trustworthy reasoning systems. By shifting evaluation beyond episodic accuracy toward bounded internal behavior under uncertainty, the framework provides a principled foundation for designing and assessing reliable hybrid reasoning systems in real-world deployment settings.

\bibliography{references}

\end{document}